\documentclass{article}
\usepackage{spconf,amsmath,graphicx,hyperref}
\usepackage{graphicx} 
\usepackage{fancyhdr}     
\usepackage{tikz}
\usepackage{stfloats}
\usepackage{amsmath}
\usepackage{amssymb}
\usepackage[ruled,linesnumbered]{algorithm2e} 
\usepackage{amsmath} 
\usepackage{xcolor}  
\usepackage{multirow}
\usepackage{textcomp}

\title{FaceSleuth-R: Adaptive Orientation-Aware Attention for Robust Micro-Expression Recognition}
\name{%
  \parbox{\textwidth}{\centering
  Linquan Wu$^{1}$\textsuperscript{*}\quad
  Tianxiang Jiang$^{2}$\textsuperscript{*}\quad
  Haoyu Yang$^{5}$\textsuperscript{*}\quad
  Wenhao Duan$^{3}$\\
  Shaochao Lin$^{6}$\quad
  Zixuan Wang$^{1}$\quad
  Jacky Keung$^{1}$\quad
  Yini Fang$^{4}$}
}
\address{%
  $^{1}$ City University of Hong Kong\quad
  $^{2}$ University of Science and Technology of China\\
  $^{3}$ Ocean University of China\quad
  $^{4}$ Hong Kong University of Science and Technology\\
  $^{5}$ University of Electronic Science and Technology of China\quad
  $^{6}$ Harbin Engineering University
}

\DeclareMathOperator{\round}{round}
\DeclareMathOperator{\clamp}{clamp}
\newcommand\blfootnote[1]{%
  \begingroup
  \renewcommand\thefootnote{}%
  \footnotetext{#1}
  \addtocounter{footnote}{-1}
  \endgroup
}
\begin{document}
%

\maketitle

\begin{figure*}[b]
    \centering
    \includegraphics[width=1\textwidth]{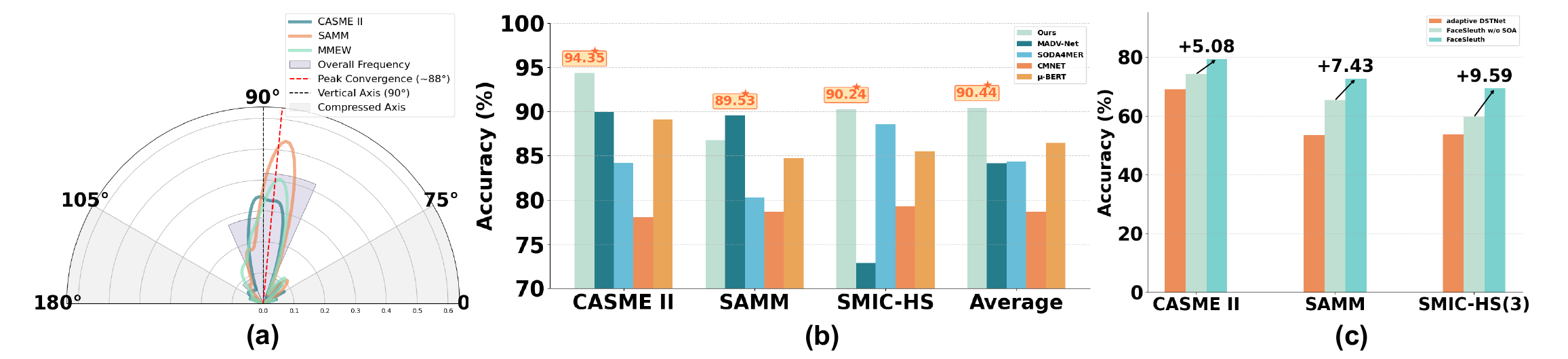} 
    \caption{Our proposed \textsc{FaceSleuth} achieves state-of-the-art performance and strong generalization ability.
    (a) The learned orientation $\theta$ consistently converges to a near-vertical prior ($\sim$88$^\circ$) across benchmarks.
    (b) Our model outperforms recent SOTA methods in intra-dataset evaluations.
    (c) On the more challenging LODO task, an ablation study further demonstrates that our key components significantly improve the model's generalization to unseen datasets.}
    \label{fig:main_fig}
\end{figure*}
\begin{abstract}
\blfootnote{\textsuperscript{*}\,Equal contribution.}
Micro-expression recognition (MER) has achieved impressive accuracy in controlled laboratory settings. However, its real-world applicability faces a significant generalization cliff, severely hindering practical deployment due to poor performance on unseen data and susceptibility to domain shifts. Existing attention mechanisms often overfit to dataset-specific appearance cues or rely on fixed spatial priors, making them fragile in diverse environments. We posit that robust MER requires focusing on quasi-invariant motion orientations inherent to micro-expressions, rather than superficial pixel-level features. To this end, we introduce \textbf{FaceSleuth-R}, a framework centered on our novel \textbf{Single-Orientation Attention (SOA)}  module. SOA is a lightweight, differentiable operator that enables the network to learn layer-specific optimal orientations, effectively guiding attention towards these robust motion cues. Through extensive experiments, we demonstrate that SOA consistently discovers a universal near-vertical motion prior across diverse datasets. More critically, FaceSleuth-R showcases superior generalization in rigorous Leave-One-Dataset-Out (LODO) protocols, significantly outperforming baselines and state-of-the-art methods when confronted with domain shifts. Furthermore, our approach establishes \textbf{state-of-the-art results} across several benchmarks. This work highlights adaptive orientation-aware attention as a key paradigm for developing truly generalized and high-performing MER systems.

\end{abstract}
%
\section{Introduction}
Micro-expressions (MEs) are involuntary, millisecond-long facial muscle activations that reveal concealed emotions \cite{davison2016samm}. Their diagnostic value has found applications in psychotherapy \cite{zhu2017comparison}, criminal investigations, and national security \cite{see2019megc}. However, MER has reached a critical juncture. While current models demonstrate high accuracy in intra-dataset evaluations, they suffer a dramatic performance decline when tested across datasets. This "generalization gulf" highlights a fundamental flaw: existing methods often overfit to dataset-specific features, failing to generalize to new domains. As a result, MER models trained in controlled settings are fragile in real-world applications, where diverse populations and environments challenge their robustness.

The fragility of these models is rooted in the reliance on spatially-bound attention mechanisms. Approaches like \textmu-BERT~\cite{nguyen2023micron} and CMNET~\cite{wei2023cmnet} focus on identifying salient spatial regions, while others such as MADV-Net~\cite{kong20253d} capture motion cues from predefined facial patches. These methods assume that the location of pixel-based features is the most reliable signal. Even models like SODA4MER~\cite{zhang2025dynamic}, which incorporate oriented deformation, still operate within spatially transformed feature maps rather than altering the directional nature of the attention itself. Such assumptions are easily disrupted by domain shifts, where lighting, subject identity, or camera perspectives change, leading to significant loss of generalization.

In this work, we propose a paradigm shift in micro-expression recognition (MER) by focusing on motion orientations rather than spatial features. We argue that intrinsic motion orientations, a bio-mechanical property inherent to facial expressions, provide a more robust and domain-invariant signal. Unlike conventional approaches that over-rely on dataset-specific spatial cues, this shift allows for more reliable performance across diverse conditions. To realize this, we introduce \textbf{FaceSleuth-R}, a framework centered on the novel \textbf{Single-Orientation Attention} (SOA) module. FaceSleuth-R employs a dual-stream architecture that effectively decouples appearance-based features from motion-based features. The SOA module operates exclusively within the motion stream, extracting directional cues while avoiding the biases introduced by dataset-specific textures.

The primary innovation of SOA lies in its ability to learn optimal attention directions in an end-to-end manner, uncovering the most generalizable motion cues directly from the data. By focusing on how the face moves rather than where it changes, SOA builds feature representations that are more resilient to domain-specific variations. Our experiments show that SOA autonomously converges to a near-vertical motion axis (~88°), aligning with a meaningful and stable physiological prior across datasets. Notably, in rigorous Leave-One-Dataset-Out (LODO) evaluations, FaceSleuth-R significantly outperforms existing state-of-the-art methods, demonstrating superior generalization capabilities.

Our contributions are threefold:
\begin{enumerate}
    \item We identify a critical bottleneck in MER: the failure of spatially-bound attention mechanisms to generalize across domains.
    \item We propose a robust paradigm that shifts the focus from spatial locations to domain-invariant directional priors of muscle motion.
    \item We introduce the Single-Orientation Attention (SOA) module, achieving state-of-the-art generalization performance in cross-dataset evaluations.
\end{enumerate}

\section{Method}
\label{sec:method}

\begin{figure}[!t] 
  \centering
  \includegraphics[width=0.5\textwidth]{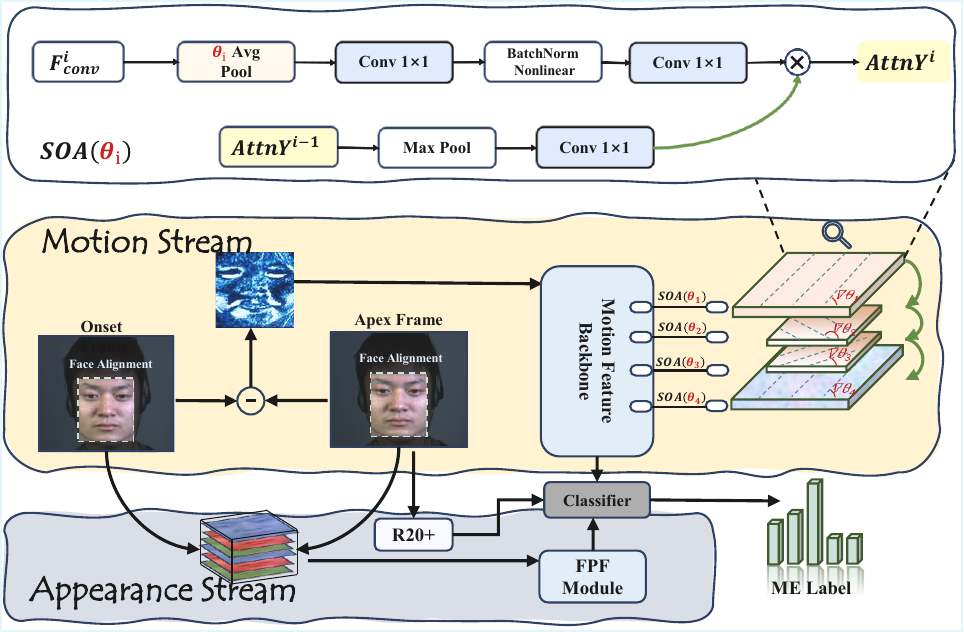} 
    \caption{FaceSleuth-R: A dual-stream framework that enhances generalization by separating motion (SOA($\theta_i$)) and appearance (FPF) cues. The features are fused and passed to a classifier to yield $\Delta_{\mathrm{final}}$.}
  \label{fig:diagram}
\end{figure}

\subsection{Motion Stream: Learning Domain-Invariant Orientations with SOA}
\label{sec:motion_stream}

The \textit{Motion Stream} is the cornerstone of FaceSleuth-R's generalization capability. Its primary function is to distill robust, domain-invariant directional cues from the input motion representation, thereby resisting the overfitting to superficial, dataset-specific artifacts that plagues conventional MER models. The central innovation within this stream is our proposed SOA module. We reposition SOA from a mere filter to a dynamic, bias-resistant mechanism that actively shifts the model's focus from \textit{where} the face changes to the more fundamental question of \textit{how} it moves.

\textbf{Conceptual Framework of SOA.}The core hypothesis of SOA is that the intrinsic orientation of muscle-driven facial movements constitutes a quasi-invariant prior that is far more resilient to domain shifts than pixel-level textures or spatial locations. To capture this prior, SOA is designed to perform two key functions: first, to learn the statistically optimal orientation $\theta$ for amplifying micro-expression signals directly from the data; and second, to re-weight the feature map to enhance motion cues along this learned direction. This process forces the network to construct a feature representation based on a more abstract and generalizable property of the motion, rather than memorizing spurious correlations present in the training domain.

\textbf{Formulation of Single-Orientation Attention.}
The SOA module maps an input feature map $F\in\mathbb{R}^{C\times H\times W}$ to an orientation-enhanced output $\tilde{F}$ via a three-stage process (see Algorithm~\ref{alg:soa} and the inset of Fig.~\ref{fig:diagram}):

(1) \textit{Orientation-Aware Pooling (OAP).}  
It replaces regular grid pooling by averaging features along parallel lines defined by a learnable orientation $d_{\theta}=(\cos\theta,\sin\theta)$ (visualized as the $\theta_i$ Avg Pool block in the inset). A binary sampling mask $M_{\theta}$ selects coordinates along each parallel line; the pooled 1D vector $v_{\theta}\in\mathbb{R}^{C\times L}$ is computed as
\begin{equation}
\label{eq:oap}
(v_{\theta})_{c,j}=\frac{1}{|\mathcal{P}_{j,\theta}|}\sum_{(x,y)\in\mathcal{P}_{j,\theta}}F_{c,x,y},
\end{equation}
where $\mathcal{P}_{j,\theta}$ indexes the coordinates selected by the $j$-th line. This collapses 2D spatial information into a direction-specific 1D representation.

(2) \textit{Attention Vector Generation.}  
The pooled vector $v_{\theta}$ is fed to a lightweight bottleneck (two $1\times1$ convolutions $f_1,f_2$ with GELU) to produce a channel-wise attention vector
\begin{equation}
\label{eq:attention_vec}
a_{\theta}=\sigma\big(f_2(\mathrm{GELU}(f_1(v_{\theta})))\big),
\end{equation}
where $\sigma$ is the sigmoid, yielding per-channel weights in $[0,1]$.

(3) \textit{Feature Re-weighting.}  
The attention $a_{\theta}$ is broadcast to spatial dimensions and applied element-wise to $F$ to produce the orientation-enhanced map
\begin{equation}
\label{eq:reweight}
\tilde{F}=a_{\theta}\otimes F.
\end{equation}
This re-weighting amplifies channels with strong directional responses while suppressing irrelevant ones, producing features attuned to generalizable motion cues.

\begin{algorithm}
\footnotesize 
\KwIn{$F$: tensor $[B, C, H, W]$, $\theta$: scalar in radians}
\KwOut{Re-weighted feature map}
\BlankLine

\SetKwFunction{FMain}{\texttt{main}} 

\SetKwProg{Fn}{Function}{:}{}
\Fn{\FMain}{
    $S \leftarrow \operatorname{int}\left(\operatorname{abs}\left(\frac{1}{\tan(\theta)}\right).\round().\clamp(\min=1\text{e-}2)\right)$\;
    
    \If{$0 < \theta < \frac{\pi}{2}$}{
        $F_{pad} \leftarrow F.\texttt{pad}((S \times (H-1), 0))$\;
    }
    \Else{
        $F_{pad} \leftarrow F.\texttt{pad}((0, S \times (H-1)))$\;
    }
    
    $lines \leftarrow \texttt{unfold\_along\_theta}(F_{pad}, \theta, \text{step}=S)$;  
    $v \leftarrow \texttt{mean}(lines, \text{dim}=2, \text{keepdim}=\text{True})$;   Eq. \ref{eq:oap}

    $w \leftarrow \texttt{torch.sigmoid}(f_2(\texttt{GELU}(f_1(v))))$;   Eq. \ref{eq:attention_vec}

    $w_{grid} \leftarrow \texttt{fold\_back}(w, \theta, H, W, \text{step}=S)$; 

    \Return{$F \odot w_{grid}$};   Eq. \ref{eq:reweight}
}
\caption{Pseudocode for a single SOA layer}
\label{alg:soa}
\end{algorithm}
\textbf{Multi-Layer Independent Attention.}
To improve the robustness and flexibility of our approach, FaceSleuth-R uses multiple independent SOA modules stacked in a multi-layer structure. Each SOA module learns its own optimal orientation, allowing the model to capture motion priors at different scales. As shown in Algorithm \ref{alg:soa}, each layer operates with a different orientation $\theta_i$, which is learned during training. This setup enables each layer to adapt its attention mechanism to capture specific directional cues at various levels.

Unlike methods that use a fixed orientation or a single global one, our independent attention layers allow for more flexible and precise feature extraction. In Algorithm \ref{alg:soa}, the orientation $\theta$ is independently adjusted for each layer to re-weight the feature maps according to the learned direction (Lines 9–11). This independent adjustment across layers allows the model to capture more complex, multi-scale motion features, improving its generalization across datasets.

\subsection{Appearance Stream and Feature Fusion}
\label{sec:appearance}
The Appearance Stream complements the Motion Stream by providing crucial spatial context, answering where movements occur to disambiguate the how learned by SOA. This stream is anchored by our Facial Position Focalizer (FPF) module, which operates on a channel-wise concatenation of the onset and apex frames. Leveraging a Swin Transformer block \cite{liu2021swin}, FPF identifies regions corresponding to key Action Units (AUs), creating a stable spatial map of facial landmarks. The final representation is formed by fusing these spatial features with the orientation-enhanced motion features via element-wise addition. This synergistic fusion is critical for generalization, as it grounds the domain-invariant motion cues from SOA with a robust spatial context from FPF, enabling more accurate interpretations on unseen faces.

\section{Experiment}
\subsection{Experimental Setup}
\label{sec:exp_setup}

\textbf{Datasets.} Our experiments are conducted on three widely-used spontaneous micro-expression datasets: CASME II~\cite{yan2014casme}, SAMM~\cite{davison2016samm}, and SMIC-HS~\cite{li2013spontaneous}. For all datasets, we use the Dlib library to perform face detection and extract 68 facial landmarks. The faces are then aligned via similarity transformation and cropped to 224 × 224 pixels. 
\textbf{Evaluation Protocols.} To comprehensively evaluate our model, we employ two distinct and rigorous protocols. 1) Intra-Dataset Evaluation: To ensure a fair comparison with previous state-of-the-art methods, we follow the standard LOSO cross-validation protocol within each dataset. This protocol assesses the model's performance on unseen subjects from the same domain. 2) Cross-Dataset Generalization: To measure the model's ability to overcome the ``generalization cliff'', we employ the more challenging LODO protocol. In this setup, a model is trained on two datasets and evaluated on the third, entirely unseen dataset, which rigorously tests its resilience to domain shifts. Importantly, CASME II and SAMM are evaluated under their native five-class emotion while SMIC-HS follows its established four-class protocol.
\textbf{Implementation Details.} Our FaceSleuth-R is built upon a ResNet-18 backbone for its motion stream. The model is trained end-to-end using the AdamW optimizer with a learning rate of $1 \times 10^{-4}$ and a batch size of 64. All experiments are implemented in PyTorch. 

\subsection{Main Results: Generalization Power and SOTA Performance}
Our primary evaluation focuses on addressing the critical "generalization cliff" in MER. The results confirm that FaceSleuth-R, by learning domain-invariant motion orientations, achieves state-of-the-art performance in the Leave-One-Dataset-Out (LODO) protocol. As detailed in Table~\ref{tab:main_results} and illustrated in Figure~\ref{fig:main_fig}(c), FaceSleuth-R consistently outperforms existing methods across multiple datasets, maintaining significantly higher accuracy with a much smaller generalization gap. This demonstrates that by focusing on how the face moves rather than where it changes, the SOA module enables our model to achieve superior generalization and resilience to domain shifts, outperforming all other methods on the LODO task.

Crucially, the strong generalization capability of FaceSleuth-R does not come at the expense of its performance on specialized, intra-dataset benchmarks. We further evaluated our model using the standard LOSO protocol, with results also presented in Table~\ref{tab:main_results} and visualized in Figure~\ref{fig:main_fig}(b). The comparison demonstrates that FaceSleuth-R establishes a new state-of-the-art, outperforming all baseline methods and ranking first in average accuracy across all benchmarks. This indicates that our insight-driven approach of focusing on fundamental motion priors not only solves the critical generalization problem but also leads to a more effective and robust feature representation for high-performance, specialized MER.

\begin{table}[t]
    \centering
    \caption{Performance comparison on Cross-Dataset and Intra-Dataset (LOSO) evaluation protocols. Best results are in \textbf{bold}. A dash (“-”) indicates that the data was not reported in the original paper.}
    \label{tab:main_results}
    \resizebox{\columnwidth}{!}{%
        \begin{tabular}{l|cc|cc|cc}
            \hline
            \multicolumn{1}{c|}{\multirow{2}{*}{\textbf{Method}}} & \multicolumn{2}{c|}{\textbf{CASME II}} & \multicolumn{2}{c|}{\textbf{SAMM}} & \multicolumn{2}{c}{\textbf{SMIC-HS}} \\
            \multicolumn{1}{c|}{} & \textbf{ACC} & \textbf{UF1} & \textbf{ACC} & \textbf{UF1} & \textbf{ACC} & \textbf{UF1} \\ \hline
            \multicolumn{7}{c}{\textbf{Cross-Dataset (LODO)}} \\ \hline
            TKRM~\cite{peng2019novel}     & 0.643 & 0.631 & 0.458 & 0.450 & 0.524 & 0.503  \\
            adptive-DSTNet~\cite{liu2025lightweight}     & 0.690 & 0.688 & 0.534 & 0.529 & 0.537 & 0.542  \\
            Meta-CDMERF~\cite{wang2025cross}     & 0.638 & 0.600 & 0.498 & 0.425 & - & - \\ 
            \textbf{FaceSleuth-R (Ours)} & \textbf{0.7429} & \textbf{0.6877} & \textbf{0.7267} & \textbf{0.7214} & \textbf{0.6922} & \textbf{0.6485} \\ \hline
            \multicolumn{7}{c}{\textbf{Intra-Dataset (LOSO)}} \\ \hline
            $\mu$-BERT~\cite{nguyen2023micron} & 0.8348 & 0.8553 & 0.8386 & 0.8475 & 0.8550 & 0.8384 \\
            MADV-Net~\cite{kong20253d}     & 0.8994 & 0.8400 & \textbf{0.8953} & \textbf{0.8800} & 0.7287 & 0.6900 \\
            AU-GCN~\cite{xie2020assisted}     & 0.7427 & 0.7040 & 0.7426 & 0.7045 & 0.6872 & 0.7012 \\
            FRL-DGT~\cite{zhai2023feature}     & 0.7570 & 0.7480 & 0.7190 & 0.7340 & 0.7460 & 0.7460  \\
            Micro-ExpMultNet~\cite{zhao2022multimodal}& 0.9034 & 0.8947 & - & - & 0.7999 & 0.7812  \\
            MMNet~\cite{li2022mmnet}     & 0.8835 & 0.8676 & 0.8014 & 0.7291 & - & -  \\
            SODA4MER~\cite{zhang2025dynamic}     & 0.8141 & 0.8418 & 0.7893 & 0.8030 & 0.8855 & 0.8881 \\ 
            \textbf{FaceSleuth-R (Ours)} & \textbf{0.9435} & \textbf{0.9402} & 0.8673 & 0.8171 & \textbf{0.9024} & \textbf{0.8983} \\ \hline
        \end{tabular}%
    }
\end{table}
\subsection{Analysis and Ablation Study}
To understand the source of FaceSleuth-R's superior generalization capabilities, we conducted a series of analyses to dissect the behavior and contribution of our core SOA module.

\begin{figure}[htbp]
    \centering
    \includegraphics[width=0.45\textwidth]{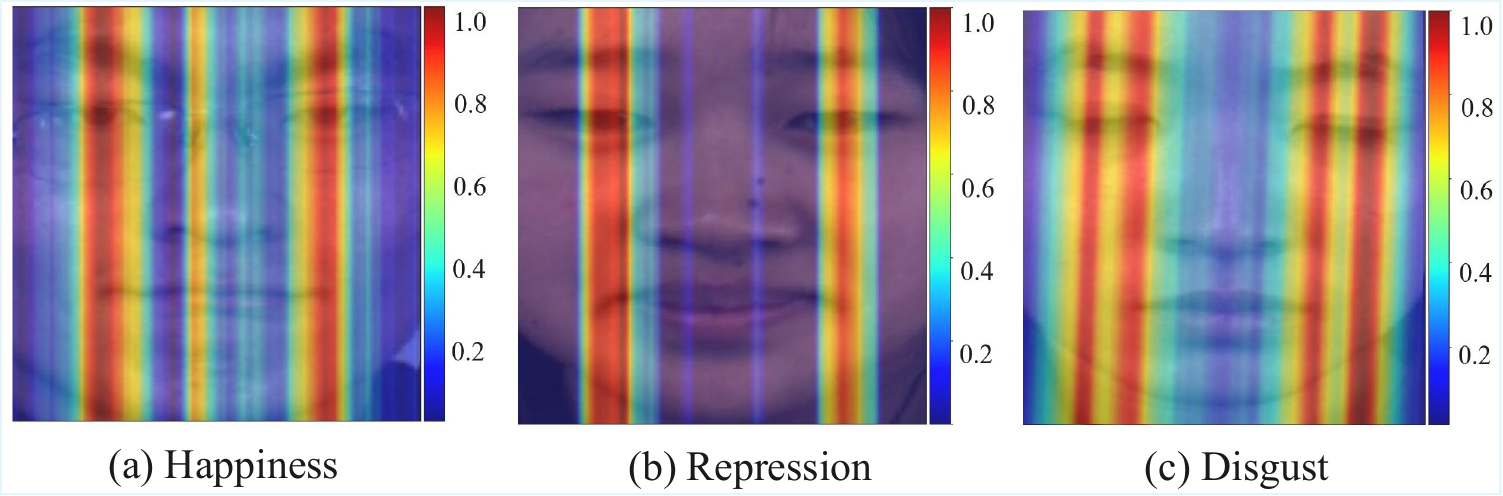}
    \caption{Visualization of attention maps generated by the SOA.}
    \label{fig:gradcam}
\end{figure}

\textbf{Learned Universal Prior.}
We first examined the learned orientation prior. As shown in Figure~\ref{fig:main_fig}(a), the orientation $\theta$ consistently converges to a near-vertical axis ($\sim$88$^\circ$) across all datasets. This confirms that SOA learns a universal, domain-invariant motion prior, not dataset-specific artifacts, demonstrating the robustness of focusing on motion orientation for generalization.

\textbf{Quantitative Impact via Ablation.}
An ablation study comparing FaceSleuth-R with and without the SOA modules (Table~\ref{tab:ablation}) reveals significant improvements. SOA boosts performance in intra-dataset (LOSO) evaluation (e.g., +3.23\% on CASME II), with an even more pronounced effect in cross-dataset (LODO) evaluation (e.g., +7.43\% on SAMM), highlighting its crucial role in enhancing generalization.

\textbf{Semantic Meaning of Orientations.}
We also examined whether the learned orientations have semantic meaning by visualizing $\theta$ for different emotions on CASME II (Figure~\ref{fig:gradcam}). The orientations varied slightly by emotion (e.g., happiness at 90.4$^\circ$ and disgust at 87.5$^\circ$), indicating that SOA captures emotion-specific motion signatures, supporting its hierarchical, nuanced representation of motion.

\begin{table}[h!]
\centering
\caption{Ablation study of the SOA module. The performance gain is significantly more pronounced in the cross-dataset (LODO) setting, highlighting its primary contribution to generalization.}
\label{tab:ablation}
\resizebox{\columnwidth}{!}{%
\begin{tabular}{l|cc|cc}
\hline
\multicolumn{1}{c|}{\multirow{2}{*}{\textbf{Method}}} & \multicolumn{2}{c|}{\textbf{Intra-Dataset (LOSO)}} & \multicolumn{2}{c}{\textbf{Cross-Dataset (LODO)}} \\
\multicolumn{1}{c|}{} & \textbf{CASME II} & \textbf{SAMM} & \textbf{CASME II} & \textbf{SAMM} \\ \hline
Baseline (w/o SOA) & 0.9112 & 0.8346 & 0.7418 & 0.6524 \\
\textbf{FaceSleuth-R (w/ SOA)} & 0.9435 & 0.8673 & 0.7926 & 0.7267  \\ \hline
\textbf{Improvement} & \textbf{+0.0323} & \textbf{+0.0327} & \textbf{+0.0508} & \textbf{+0.0743} \\ \hline
\end{tabular}%
}
\end{table}

\section{Conclusion}
\label{sec:conclusion}

FaceSleuth-R addresses MER generalization by prioritizing motion orientations over spatial features via its SOA module. It identifies a universal near-vertical motion prior, significantly improving LODO performance and setting new intra-dataset benchmarks. The model achieves high accuracy and robust generalization, highlighting the importance of orientation-aware attention. Future work will test its robustness in real-world scenarios such as facial occlusions.

\bibliographystyle{IEEEbib}
\bibliography{strings,refs}

\end{document}